# Benchmarking Large Multimodal Models for Ophthalmic Visual Question Answering with OphthalWeChat

## Running Title: OphthalWeChat: Multimodal VQA Benchmark

Pusheng Xu, MD[1#], Xia Gong, MD[1#], Xiaolan Chen, MD[1], Weiyi Zhang, MS[1], Jiancheng Yang, PhD[2], Bingjie Yan, MS[1], Meng Yuan, MD, PhD[3], Yalin Zheng, MEng, PhD[4], Mingguang He, MD, PhD[1,5,6], Danli Shi, MD, PhD[1,5*]

**Affiliations**

1. School of Optometry, The Hong Kong Polytechnic University, Kowloon, Hong Kong.
2. Swiss Federal Institute of Technology Lausanne (EPFL), Lausanne, Switzerland.
3. Wuhan Research Center, PolyU-Wuhan Technology and Innovation Research Institute, Wuhan, Hubei, China.
4. Department of Eye and Vision Sciences, University of Liverpool, Liverpool, United Kingdom.
5. Research Centre for SHARP Vision, The Hong Kong Polytechnic University, Kowloon, Hong Kong.
6. Centre for Eye and Vision Research (CEVR), 17W Hong Kong Science Park, Hong Kong

# Contributed equally

**Correspondence**

***Dr Danli Shi,** MD, PhD., Experimental Ophthalmology, School of Optometry, The Hong Kong Polytechnic University, Kowloon, Hong Kong.

**Email:** danli.shi@polyu.edu.hk

**Abstract**

**Importance:** Vision–language models (VLMs) hold promise for interpreting ophthalmic images, but there lacks a standardized benchmark to assess their performance across multimodal imaging and subspecialties.

**Objective:** To develop a bilingual multimodal visual question answering (VQA) benchmark for evaluating VLMs in ophthalmology.

**Design, Setting, and Participants**: Ophthalmic image posts and associated captions published between January 1, 2016, and December 31, 2024, were collected from WeChat Official Accounts. Based on these captions, bilingual question–answer (QA) pairs in Chinese and English were generated using GPT-4o-mini. QA pairs were categorized into six subsets by question type and language: binary (Binary_CN, Binary_EN), single-choice (Single-choice_CN, Single-choice_EN), and open-ended (Open-ended_CN, Open-ended_EN). The benchmark was used to evaluate the performance of three VLMs: GPT-4o, Gemini 2.0 Flash, and Qwen2.5-VL-72B-Instruct.

**Exposures:** VLMs were prompted with images and corresponding questions to generate answers.

**Main Outcomes and Measures:** The primary outcome was overall accuracy for each model. Secondary outcomes included subset-specific accuracy and open-ended question performance using BLEU-1 and BERTScore metrics.

**Results:** The final OphthalWeChat dataset included 3,469 images and 30,120 QA pairs across 9 ophthalmic subspecialties, 548 conditions, 29 imaging modalities, and 68 modality combinations. Gemini 2.0 Flash achieved the highest overall accuracy (0.548), outperforming GPT-4o (0.522, $P < 0.001$) and Qwen2.5-VL-72B-Instruct (0.514, $P < 0.001$). It also led in both Chinese (0.546) and English subsets (0.550). Subset-specific performance showed Gemini 2.0 Flash excelled in Binary_CN (0.687), Single-choice_CN (0.666), and Single-choice_EN (0.646), while GPT-4o ranked highest in Binary_EN (0.717), Open-ended_CN (BLEU-1: 0.301; BERTScore: 0.382), and Open-ended_EN (BLEU-1: 0.183; BERTScore: 0.240).

**Conclusions and Relevance:** This study presents the first bilingual VQA benchmark for ophthalmology, distinguished by its real-world context and inclusion of multiple examinations per patient. The dataset reflects authentic clinical decision-making scenarios and enables quantitative evaluation of VLMs, supporting the development of accurate, specialized, and trustworthy AI systems for eye care.

## INTRODUCTION

Recent advances in large language models (LLMs) have significantly impacted healthcare by enabling models to make clinically relevant judgments with strong reasoning capabilities.[1,2] However, traditional LLMs lack the ability to integrate multimodal information. Vision–language models (VLMs) and multimodal LLMs address this limitation by jointly processing visual and textual inputs, thereby unlocking capabilities that surpass those of single-modal LLMs, particularly in fields requiring integrated cross-modal reasoning.[3,4] In ophthalmology, multimodal LLMs and agents have shown considerable potential, especially in tasks such as Visual Question Answering (VQA), where models interpret ophthalmic images and provide responses to clinically meaningful questions.[5-11] However, the development and automated evaluation of VLMs rely on access to diverse, high-quality datasets and standardized benchmarks to ensure robust performance comparison and validation.[12]

Despite growing interest, publicly available ophthalmic datasets remain scarce due to data access constraints. Although recent initiatives have produced open datasets for tasks including disease diagnosis, image segmentation, report generation, and treatment response prediction,[13-15] these resources often lack sufficient diversity in imaging modalities, clinical complexity, and annotation granularity, making them inadequate for generating conversationally rich question–answer (QA) pairs. To address these limitations, some researchers have begun leveraging publicly available content from social media platforms to build VQA datasets and train VLMs. For instance, the OpenPath dataset was curated by mining medical images and captions from Twitter, illustrating the feasibility and scalability of sourcing VQA-relevant data from online public repositories.[16]

Ophthalmology is a highly image-centric specialty that utilizes a wide array of imaging modalities, including color fundus photography (CFP), optical coherence tomography (OCT), and fundus fluorescein angiography (FFA), among others. Ocular imaging plays a crucial role not only in diagnosing eye diseases but also in revealing systemic conditions, as the transparent ocular media provide direct visualization of intraocular structures, neural tissues, and microvasculature. Yet, existing medical VQA benchmarks, such as VQA-RAD,[17] PathVQA,[18] DermaVQA,[19] and PMC-VQA,[20] focus primarily on other specialties and offer little or no ophthalmic content. Furthermore, the increasing demand for multilingual clinical AI systems remains largely unmet in current VQA resources.[21]

To address these gaps, we developed OphthalWeChat, a large-scale, bilingual (Chinese–English) VQA benchmark specifically designed for ophthalmology. This resource aims to support the development and rigorous evaluation of VLMs in realistic ophthalmic scenarios, enabling applications in automated diagnosis, medical education, and telemedicine.

## METHODS

The construction process of the OphthalWeChat is shown in **Figure 1** and includes five key steps: (1) collection of ophthalmic image posts from WeChat, (2) extraction and classification of image–caption pairs, (3) generation of bilingual (Chinese–English) question–answer (QA) pairs, (4) quality control and answer balance adjustment, and (5) finalization of the dataset and model evaluation.

### Data collection

Following a methodology similar to that used in a prior study published in *Nature Medicine*,[16] which collected pathology images from social media, we curated ophthalmic images and their corresponding captions from WeChat Official Accounts. WeChat is a publicly accessible platform that allows verified users—such as healthcare professionals and institutions—to disseminate text- and image-based medical content.

For this study, we included posts published between January 1, 2016, and December 31, 2024. In accordance with WeChat's content policies and applicable legal and ethical guidelines, all data included in the dataset are traceable to their original sources.

This research did not require Institutional Review Board approval or informed consent, as it does not involve data collection from human subjects. All ophthalmic content was publicly available and anonymized, with no personally identifiable or sensitive information included at the time of original publication.

**Data processing**

Ophthalmic images and their corresponding captions were systematically extracted from the collected WeChat posts using custom Python scripts. Images comprising multiple sub-figures under a shared caption were retained as composite figures and were not further segmented.

Following extraction, non-ophthalmic images and irrelevant textual content were manually excluded by an ophthalmologist with over six years of clinical experience. To prevent models from leveraging Optical Character Recognition (OCR) to retrieve diagnostic information directly, any textual annotations within images containing diagnoses were cropped out.

Each image was categorized into one of nine ophthalmic subspecialties, following the classification framework of the American Academy of Ophthalmology (AAO) (https://www.aao.org). Imaging modality information was extracted from the corresponding captions. When modality details were not explicitly stated, classification was manually determined by the ophthalmologist based on visual features and clinical context.

**QA pairs generation**

Chinese-English bilingual QA pairs, including close-ended formats (binary and single-choice questions with four answer options) and open-ended questions, were generated using GPT-4o-mini between January 14 and 16, 2025, using prompt listed in **Supplementary Table 1**.

Due to the heterogeneous nature of the image captions, which ranged from concise diagnostic labels to detailed multi-image descriptions, no fixed upper limit was set for the number of QA pairs derived per caption. This approach ensured flexibility and allowed for comprehensive coverage of clinically relevant content across diverse ophthalmic scenarios.

**Quality control and label balance adjustment**

To ensure that each QA pair could be answered based solely on the visual content of the corresponding image, the initial prompt emphasized lesion- and diagnosis-related questions. Nevertheless, some generated QA pairs included information not inferable from the image alone, such as visual acuity, intraocular pressure, symptoms, medical history, or treatment details. These QA pairs were systematically excluded through keyword-based filtering, using an exclusion list provided in **Supplementary Table 2.**

To mitigate potential bias due to imbalanced answer distributions in binary and single-choice QA formats, corrective adjustments were implemented. Among binary QA pairs, the initial distribution was skewed, with 71.2% labeled as "True" and 28.8% as "False." To address this, 21.2% of the “True” binary questions and their corresponding captions were randomly selected and regenerated into semantically opposite forms, now labeled as “False”, using the prompt outlined in **Supplementary Table 3A**.

A similar imbalance was observed in single-choice QA pairs, where the correct answer distribution was: A (39.2%), B (44.0%), C (14.7%), and D (2.1%). To achieve a uniform distribution (25% per option), QA pairs originally labeled as A or B were randomly sampled and modified by reassigning the correct answers to C or D through controlled option-swapping, guided by the procedure in **Supplementary Table 3B**.

The finalized dataset was stratified into six subsets based on question type and language: binary QA in Chinese (Binary_CN) and English (Binary_EN); single-choice QA in Chinese (Single-choice_CN) and English (Single-choice_EN); and open-ended QA in Chinese (Open-ended_CN) and English (Open-ended_EN).

**VLMs performance evaluation**

We evaluated the performance of three representative VLMs: GPT-4o (version: 2024-11-20, OpenAI), Gemini 2.0 Flash (Google DeepMind), and Qwen2.5-VL-72B-Instruct (Alibaba DAMO Academy). These models were selected based on their public accessibility, recent release dates, robust multimodal capabilities, and demonstrated utility in both academic and industrial applications. Notably, the selected models vary in architecture scale, training paradigms, and geographic origin. All evaluations were conducted via official Application Programming Interfaces (APIs), and responses were generated in the same language as the corresponding question, following prompt structures detailed in **Supplementary Table 4**. The evaluation spanned from January 17 to February 14, 2025.

Model performance on both closed-ended and open-ended VQA tasks was primarily assessed using accuracy. For open-ended responses, we additionally computed BLEU-1 scores[22] to evaluate lexical overlap and BERTScore[23] to assess semantic similarity with reference answers.

To assess the factual correctness of open-ended responses, we employed DeepSeek-V3 via API, using a standardized evaluation prompt that included the input question, reference answer, and model-generated response (see **Supplementary Table 5**). To verify the reliability of DeepSeek-V3 as an automated evaluator, we conducted a preliminary validation on a subset of 100 open-ended QA pairs. Each output from DeepSeek-V3 was manually reviewed by the same experienced ophthalmologist previously referenced. The model demonstrated 100% concordance with expert judgment, supporting its use as a valid and trustworthy evaluator in this study.

**Statistical analysis**

All statistical analyses were conducted using Stata (version 18.0; StataCorp LLC, College Station, TX, USA). To compare accuracy between Chinese and English subsets for the same model, the chi-square test was applied. Within each subset, the model with the highest overall accuracy was designated as the reference model. Comparisons between this reference model and other models in the same subset were performed using McNemar's test for paired proportions. A two-sided P value < 0.05 was considered statistically significant.

Data visualizations, including pie charts, bar charts, and radar charts, were created using Origin 2025 (OriginLab Corporation, Northampton, MA, USA). The radial tree diagram was generated using R (version 4.3.1; R Foundation for Statistical Computing, Vienna, Austria).

## RESULTS

### OphthalWeChat dataset overview

The dataset comprises 3,469 ophthalmic images and 30,120 question–answer (QA) pairs, encompassing 548 distinct eye conditions. As illustrated in **Figure 2A,** the distribution of images across ophthalmic subspecialties is as follows: Retina/Vitreous (68.7%), Cornea/External Disease (9.6%), Cataract/Anterior Segment (7.2%), Uveitis (3.5%), Neuro-Ophthalmology/Orbit (3.1%), Glaucoma (2.3%), Ocular Pathology/Oncology (2.3%), Pediatric Ophthalmology/Strabismus (1.7%), and Oculoplastics/Orbit (1.6%). No images were classified under Comprehensive Ophthalmology or Refractive Management/Intervention.

The five most frequently represented conditions in the dataset are retinal detachment, diabetic retinopathy, macular hole, branch retinal vein occlusion, and cataract (**Figure 2B**).

The dataset includes images from 29 distinct imaging modalities and 68 modality combinations (**Supplementary Table 6**). **Figure 2C** presents the frequency distribution of the top 15 most common modalities and combinations. The five most prevalent modalities are: CFP (689 images, 19.9%), scanning laser ophthalmoscopy (SLO; 652 images, 18.8%), OCT (645 images, 18.6%), slit-lamp photography (550 images, 15.9%), and FFA (130 images, 3.75%). **Figures 2D** displays a word cloud illustrating the most frequently occurring ophthalmological terms, including "retinal," "detachment," "macular," "degeneration," and "retinopathy." **Figure 3A** showcases representative images for each imaging modality. **Figure 3B** provides examples of the generated QA pairs from the six subsets.

### VLMs overall performance

As summarized in **Table 1**, Gemini 2.0 Flash demonstrated the highest overall accuracy (0.548, reference) among the evaluated visual language models (VLMs), outperforming GPT-4o (0.522, $P < 0.001$) and Qwen2.5-VL-72B-Instruct (0.514, $P < 0.001$). This trend held across both language subsets: For Chinese, Gemini 2.0 Flash (0.546, reference) vs. GPT-4o (0.508, P <

0.001) and Qwen2.5-VL-72B-Instruct (0.519, P < 0.001); For English, Gemini 2.0 Flash (0.550, reference) vs. GPT-4o (0.537, P = 0.011) and Qwen2.5-VL-72B-Instruct (0.503, P < 0.001). While GPT-4o showed significantly better performance in English subsets compared to Chinese (P < 0.001), both Gemini 2.0 Flash (P = 0.437) and Qwen2.5-VL-72B-Instruct (P = 0.062) exhibited comparable accuracies across languages.

In the closed-ended tasks, Gemini 2.0 Flash achieved the highest accuracy in: 1. Binary_CN: 0.687 (vs. GPT-4o: 0.617, P < 0.001; Qwen2.5: 0.666, P = 0.014); 2. Single-choice_CN: 0.666 (vs. GPT-4o: 0.618, P < 0.001; Qwen2.5: 0.617, P < 0.001); 3. Single-choice_EN: 0.646 (vs. GPT-4o: 0.578, P < 0.001; Qwen2.5: 0.597, P < 0.001). However, GPT-4o led in Binary_EN (0.717, reference) compared to Gemini 2.0 Flash (0.704, P = 0.106) and Qwen2.5-VL-72B-Instruct (0.660, P < 0.001).

In the open-ended tasks, GPT-4o delivered the best performance: 1. Open-ended_CN: 0.241 (reference) vs. Gemini 2.0 Flash: 0.225 (P = 0.066); Qwen2.5: 0.215 (P < 0.001); 2. Open-ended_EN: 0.250 (reference) vs. Gemini 2.0 Flash: 0.239 (P = 0.207); Qwen2.5: 0.207 (P < 0.001). Additionally, GPT-4o achieved the highest BLEU-1 and BERTScore in both language subsets: 1. Open-ended_CN: BLEU-1 = 0.301, BERTScore = 0.382; 2. Open-ended_EN: BLEU-1 = 0.183, BERTScore = 0.240. Corresponding details are provided in **Table 1C and 1D**.

**VLMs performance across ophthalmic subspecialties**

We further evaluated the performance of the three VLMs across various ophthalmic subspecialties (**Figure 4**). Gemini 2.0 Flash achieved the highest accuracy in Ocular Pathology/Oncology across all six evaluation subsets.

For the closed-ended tasks, all models performed well in the Cornea/External Disease category, with accuracies ≥ 0.638. However, the subspecialty with the lowest performance varied among models and subsets, indicating domain-specific variability.

In the open-ended tasks, all models achieved relatively higher accuracy in the Glaucoma category (all accuracies ≥ 0.252). In contrast, Oculoplastics/Orbit and Pediatric Ophthalmology/Strabismus consistently demonstrated lower performance across all models (accuracies ≤ 0.161), suggesting greater challenges in generating accurate responses in these subspecialties.

**VLMs performance across imaging modalities**

In the four closed-ended subsets, all models achieved higher accuracy when interpreting slit-lamp images, with accuracies ≥ 0.646 (**Supplementary Figure 1**). In contrast, performance was relatively lower for ultrasound B-scan images, where all model accuracies were ≤ 0.688.

For the two open-ended subsets, the highest performance was observed with composite images combining CFP and fundus autofluorescence (FAF) (all accuracies ≥ 0.254). Conversely, accuracy was lowest for composite images incorporating SLO and fluorescein fundus angiography (FFA), with all models showing accuracies ≤ 0.121.

**DISCUSSION**

In this study, we developed OphthalWeChat, a comprehensive ophthalmic multimodal VQA benchmark designed to facilitate the training and evaluation of VLMs in realistic ophthalmic scenarios. The dataset comprises 3,469 images and 30,120 QA pairs spanning nine ophthalmic subspecialties, 29 imaging modalities, and 68 modality combinations. Notably, OphthalWeChat is the first bilingual (Chinese–English) benchmark in ophthalmology, thereby serving a significant portion of the global population, approximately one-third, who speak either Chinese or English.[24] Among the VLMs evaluated, Gemini 2.0 Flash demonstrated the highest overall accuracy.

OphthalWeChat represents one of the most comprehensive multimodal VQA benchmarks in ophthalmology to date. While previous efforts have aimed to develop ophthalmic VQA datasets, most have been limited by narrow modality coverage and the absence of open-ended questions (**Supplementary Table 7**). For example, FunBench assesses VLMs' capabilities in modality perception, anatomical understanding, lesion analysis, and disease diagnosis but includes only three modalities (CFP, OCT, and ultrawide-field fundus photography) and lacks open-ended queries.[25] Liang et al. introduced a dataset with 439 CFP and 75 OCT images, restricted to close-ended tasks.[26] OphthalVQA contains six modalities but only 60 images and 600 open-ended QA pairs.[27] The Diabetic Macular Edema VQA dataset includes only CFP images with binary questions.[28] GMAI-MMBench spans 284 datasets and 38 image modalities but includes just six ophthalmic modalities and no open-ended questions.[29] OmniMedVQA offers data across over 20 anatomical regions, yet ophthalmic data is limited to CFP and OCT.[30] PMC-VQA, a large-scale dataset with 227,000 QA pairs over 149,000 images, does not specify the quantity or diversity of ophthalmic images and supports only a single language.[20]

In contrast, OphthalWeChat offers a significantly broader scope, encompassing 29 imaging modalities and 68 modality combinations across nine ophthalmic subspecialties. It is also the first bilingual (Chinese–English) benchmark in ophthalmology to include both close- and open-ended QA formats. These features make it a uniquely practical and clinically relevant resource for evaluating VLMs in real-world ophthalmic contexts.

Among the VLMs assessed, Gemini 2.0 Flash achieved the highest overall accuracy, while GPT-4o performed better on English-language and open-ended tasks. The specific reasons behind these performance discrepancies remain unclear and warrant further investigation. Notably, all models achieved overall accuracies below 0.6, underscoring the insufficiency of current VLM performance for direct clinical deployment. This highlights the urgent need for a rigorous benchmark like OphthalWeChat to drive model improvement and support safe, effective AI-assisted ophthalmic care.

Performance varied significantly across subspecialties and modalities. All models performed poorly in Oculoplastics/Orbit and Pediatric Ophthalmology/Strabismus on open-ended questions, likely due to limited training data in these areas. In terms of imaging, the lowest accuracies in closed-ended tasks were observed for Ultrasound B-scan, and in open-ended tasks for composite images combining SLO and FFA. This may reflect the scarcity of large-scale, multimodal datasets, particularly for Ultrasound B-scan, and the complexity introduced by multi-panel images, which are common in clinical settings but underrepresented in training datasets.

Evaluating the correctness of open-ended responses remains a major challenge, as such responses exhibit high variability in language and reasoning. Unlike close-ended tasks, they lack predefined answers, complicating standardized scoring. Ensuring objectivity, consistency, and alignment with expert judgment is key. Recent studies have explored the use of LLMs as evaluators. For instance, Badshah et al. showed that ensembles of LLMs improve the reliability of open-ended QA assessment[31], while Lu et al. proposed an Analytic Hierarchy Process (AHP)-guided LLM framework to enhance human-aligned evaluation.[32] Building on these advancements, our study employs a third-party LLM as an external evaluator, enabling objective, reproducible, and scalable comparison across models while minimizing human-induced variability.

To further assess model output quality, we employed BLEU-1 for lexical similarity and BERTScore for semantic alignment. Interestingly, although Gemini 2.0 Flash ranked second in open-ended accuracy, it recorded the lowest BLEU-1 score, suggesting a disconnect between correctness and surface-level text similarity. This may result from the natural variability in language or from the fact that models were not fine-tuned on the benchmark dataset. These findings underscore the limitations of relying solely on automated metrics and emphasize the need for comprehensive evaluations integrating both human judgment and diverse metrics to ensure clinically meaningful assessments.[33]

This study has several limitations. First, the QA content is largely focused on lesion and diagnosis-related questions. Nonetheless, the benchmark's strength lies in its wide coverage of subspecialties and modalities, supporting generalizability. Future VQA datasets should expand to cover broader clinical decision-making tasks, including next-step diagnostics, treatment planning, and prognostic reasoning. Second, composite images were not decomposed into individual sub-figures. While this reflects real-world image presentation formats, it may introduce visual complexity that challenges multimodal reasoning. Implementing automated sub-figure detection and segmentation pipelines could support more granular and interpretable model evaluation.

## CONCLUSIONS

This study presents OphthalWeChat, a novel and comprehensive multimodal VQA benchmark in ophthalmology. By integrating diverse ophthalmic subspecialties, bilingual questions, and real-world clinical image–text pairs, OphthalWeChat fills a critical gap in current VQA resources. This benchmark is constructed to advance the development and evaluation of VLMs tailored to ophthalmic applications, offering a valuable foundation for building more specialized, accurate, and trustworthy AI systems in eye care.

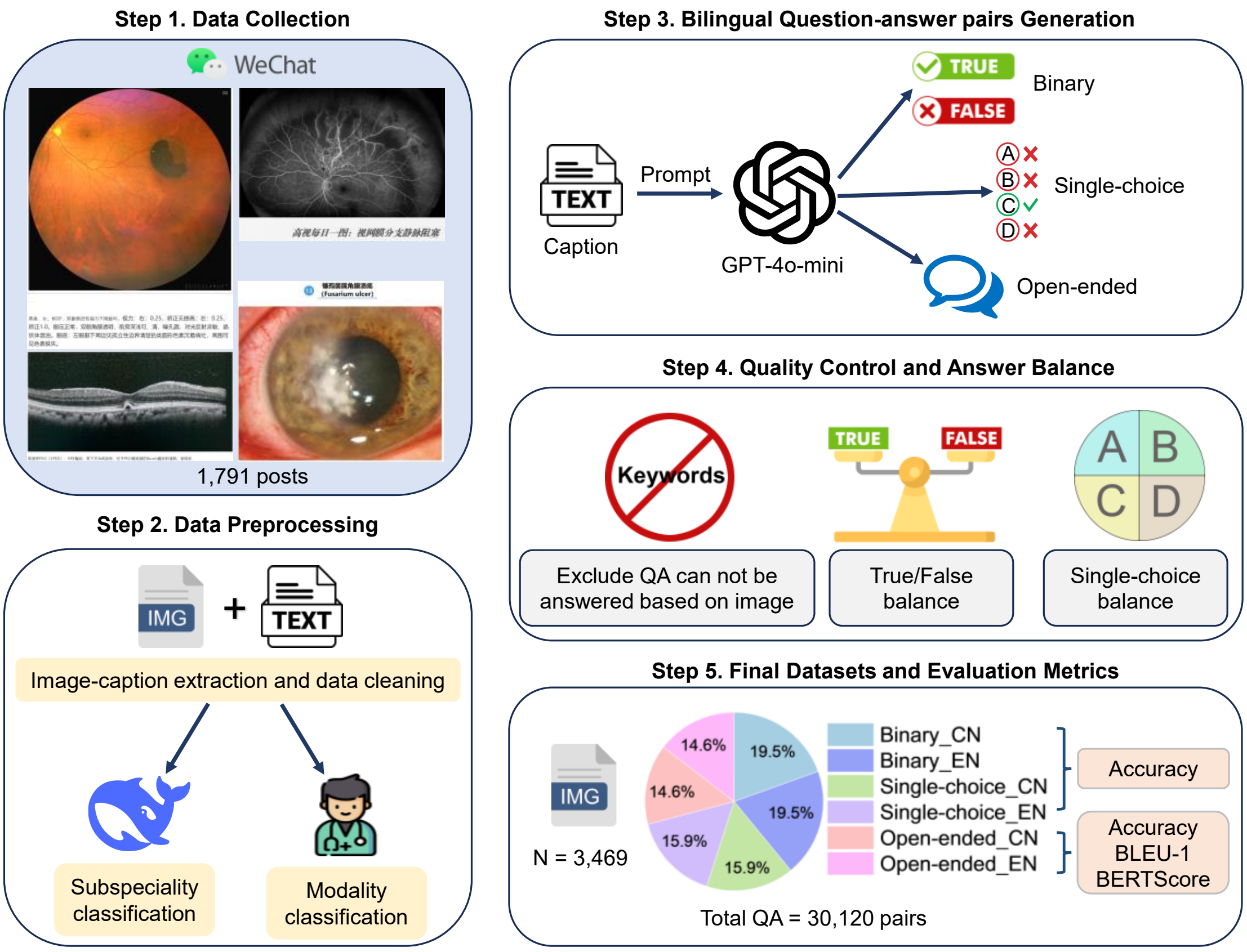

**Figure 1. Overview of the OphthalWeChat dataset construction pipeline.** The workflow includes five main stages: **(1)** collection of ophthalmic examination images from publicly available WeChat posts, **(2)** extracting ophthalmic images and their corresponding captions from the above posts, and categorize them by subspecialty and modality, **(3)** generation of Chinese–English bilingual question-answer (QA) pairs, **(4)** quality control and answer balance adjustment, and **(5)** finalization of the dataset and design of evaluation metrics. Binary_CN = binary QA pairs in Chinese; Binary_EN = binary QA pairs in English; Single-choice_CN = single-choice QA pairs in Chinese; Single-choice_EN = single-choice QA pairs in English; Open-ended_CN = open-ended QA pairs in Chinese; Open-ended_EN = open-ended QA pairs in English.

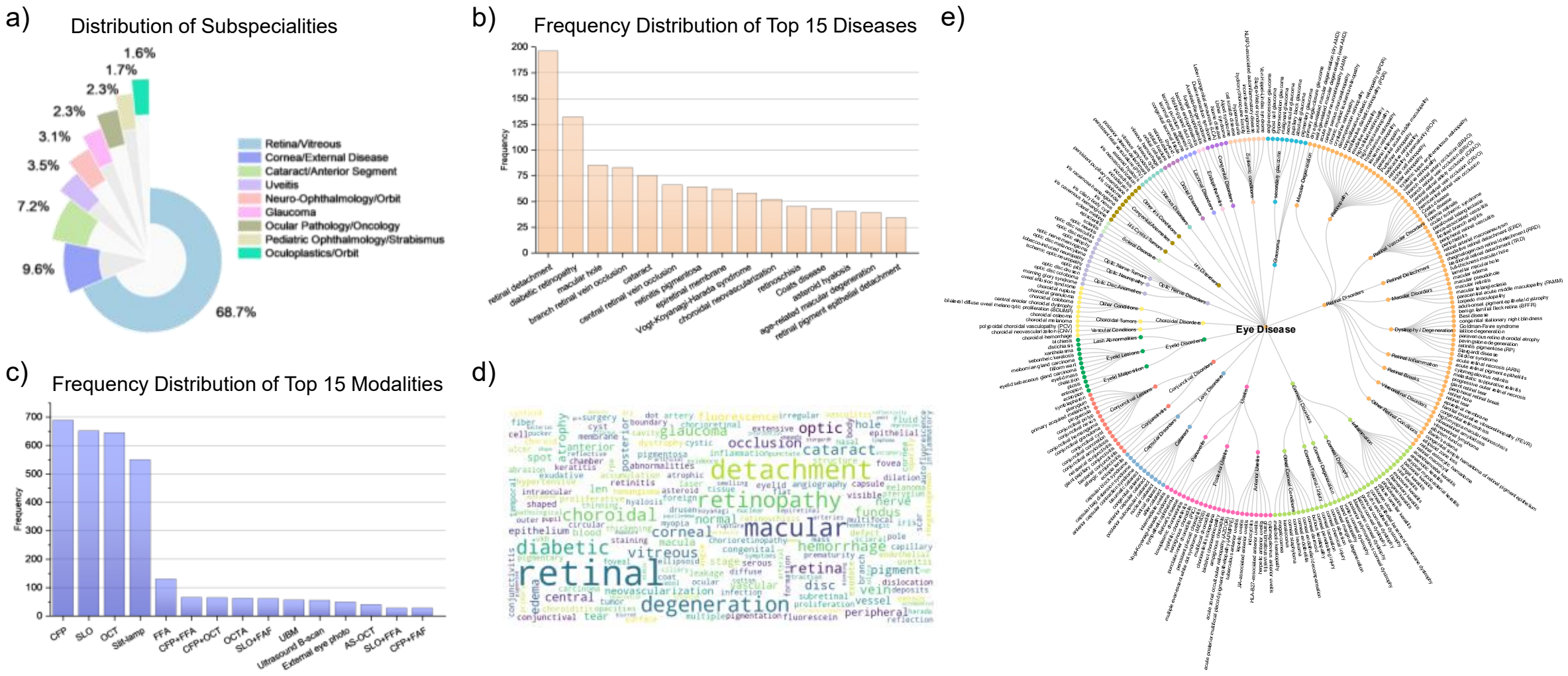

**Figure 2. Overview of the OphthalWeChat dataset. (a)** Distribution of annotated images across ophthalmic subspecialties. **(b)** Frequency distribution of the 15 most common eye diseases represented in the dataset. **(c)** Frequency distribution of the 15 most common imaging modalities and modality combinations in the OphthalWeChat dataset. Both single- and multi-modality categories are included, reflecting the multimodal nature of real-world clinical imaging. **(d)** Word cloud visualizing the most frequently occurring ophthalmologic terms in the dataset. **(e)** Radial tree diagram illustrating the structured domain knowledge hierarchy. AS-OCT = anterior segment optical coherence tomography, CFP = color fundus photography, FAF = fundus autofluorescence, FFA = fundus fluorescein angiography, OCT = optical coherence tomography, OCTA = optical coherence tomography angiography, SLO = scanning laser ophthalmoscopy, UBM = ultrasound biomicroscopy.

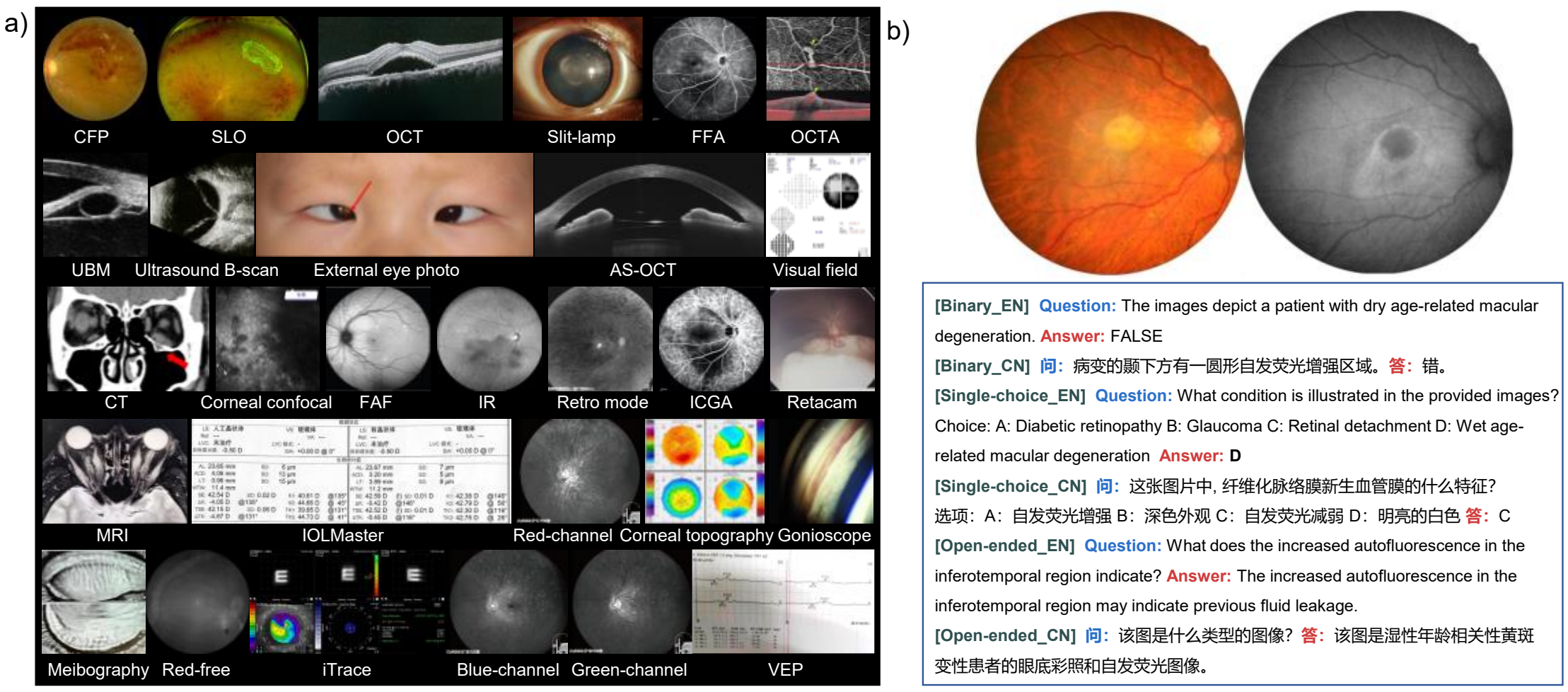

**Figure 3. Representative examples from OphthalWeChat. (a)** Sample images representing each imaging modality. **(b)** Example question-answer pairs from the six subsets, demonstrating the diversity in question types and clinical content. AS-OCT = anterior segment optical coherence tomography, CFP = color fundus photography, CT = computed tomography, FAF = fundus autofluorescence, FFA = fundus fluorescein angiography, ICGA = indocyanine green angiography, IR = infrared, MRI = magnetic resonance imaging, OCT = optical coherence tomography, OCTA = optical coherence tomography angiography, SLO = scanning laser ophthalmoscopy, UBM = ultrasound biomicroscopy, VEP = visual evoked potential.

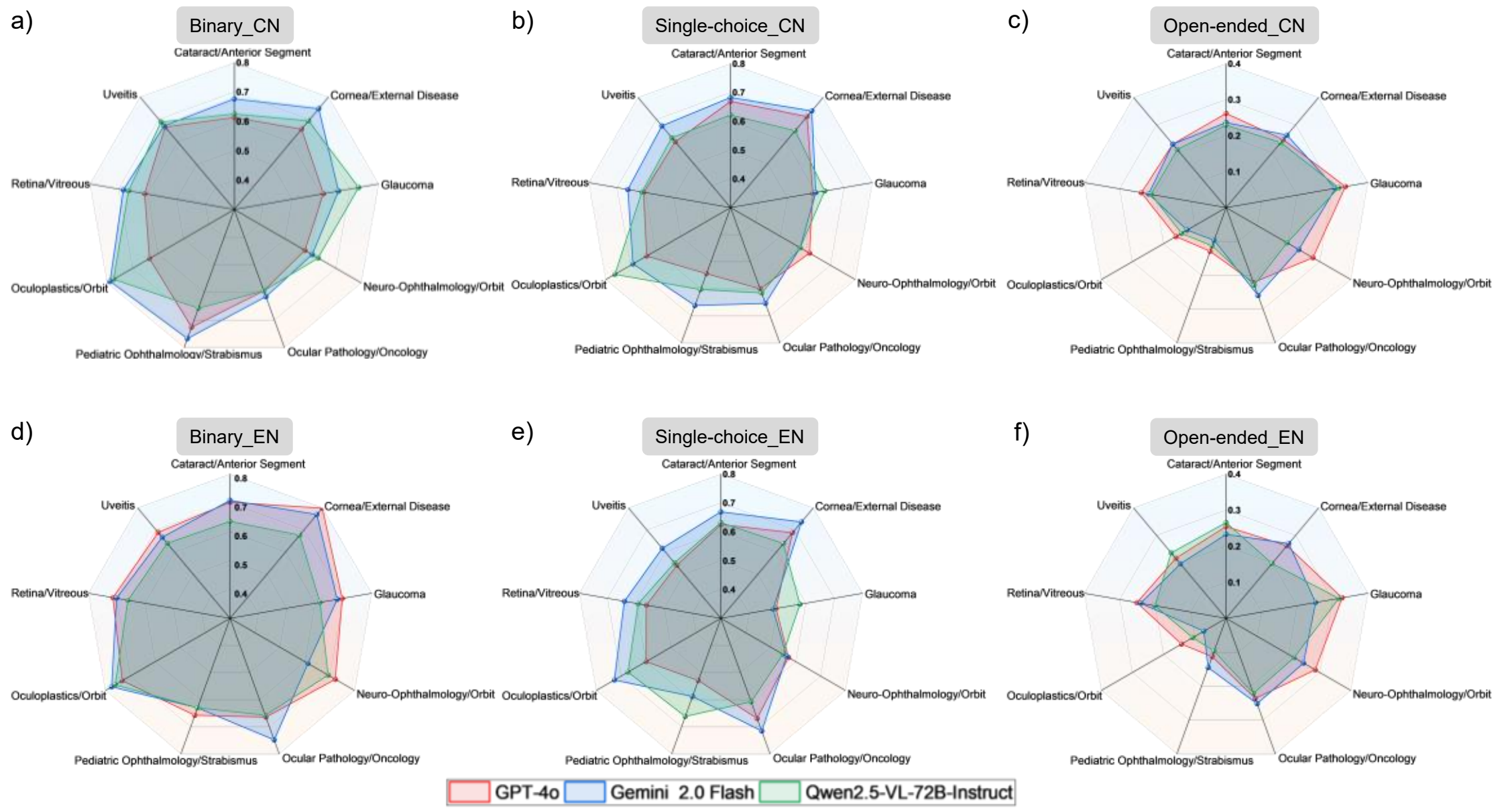

**Figure 4. Radar charts comparing the accuracy of three vision–language models across different question types and ophthalmic subspecialties.** Panels **(a–f)** display performance results for six subsets: **(a)** Binary question-answer (QA) pairs in Chinese (Binary_CN), **(b)** Single-choice QA pairs in Chinese (Single-choice_CN), **(c)** Open-ended QA pairs in Chinese (Open-ended_CN), **(d)** Binary QA pairs in English (Binary_EN), **(e)** Single-choice QA pairs in English (Single-choice_EN), and **(f)** Open-ended QA pairs in English (Open-ended_EN).

**Table 1 | Performance comparison of vision–language models on OphthalWeChat**

Panel A. Overall accuracy across all tasks (N = 30,120)

| Model | Accuracy | | | P value* |
|---|---|---|---|---|
| | Overall | Chinese | English | |
| GPT-4o | 0.522 (P<0.001) | 0.508 (P<0.001) | 0.537 (P=0.011) | < 0.001 |
| Gemini 2.0 Flash | **0.548 (Ref)** † | **0.546 (Ref)** | **0.550 (Ref)** | 0.431 |
| Qwen2.5-VL-72B-Instruct | 0.514 (P<0.001) | 0.519 (P<0.001) | 0.503 (P<0.001) | 0.062 |

Panel B. Accuracy in close-ended tasks (N = 21,350)

| Model | Binary_CN | Binary_EN | Single-choice_CN | Single-choice_EN |
|---|---|---|---|---|
| GPT-4o | 0.617 (P<0.001) | **0.717 (Ref)** | 0.618 (P<0.001) | 0.578 (P<0.001) |
| Gemini 2.0 Flash | **0.687 (Ref)** | 0.704 (P=0.106) | **0.666 (Ref)** | **0.646 (Ref)** |
| Qwen2.5-VL-72B-Instruct | 0.666 (P=0.014) | 0.660 (P<0.001) | 0.617 (P<0.001) | 0.597 (P<0.001) |

Panel C. Performance on Open-Ended_CN subset (N = 4,385)

| Model | Accuracy | BLEU-1 | BERTScore |
|---|---|---|---|
| GPT-4o | **0.241 (Ref)** | **0.301** | **0.382** |
| Gemini 2.0 Flash | 0.225 (P=0.066) | 0.229 | 0.376 |
| Qwen2.5-VL-72B-Instruct | 0.215 (P<0.001) | 0.245 | 0.325 |

Panel D. Performance on Open-Ended_EN subset (N = 4,385)

| Model | Accuracy | BLEU-1 | BERTScore |
|---|---|---|---|
| GPT-4o | **0.250 (Ref)** | **0.183** | **0.240** |
| Gemini 2.0 Flash | 0.239 (P=0.207) | 0.066 | 0.208 |
| Qwen2.5-VL-72B-Instruct | 0.207 (P<0.001) | 0.121 | 0.145 |

*P value compares the same model's performance between Chinese and English subsets using the chi-square test. †Bold values indicate the highest accuracy or score per task. P values in parentheses compare each model with the reference model (Ref) using McNemar's test. P < 0.05 was considered statistically significant.

# Supplementary Information

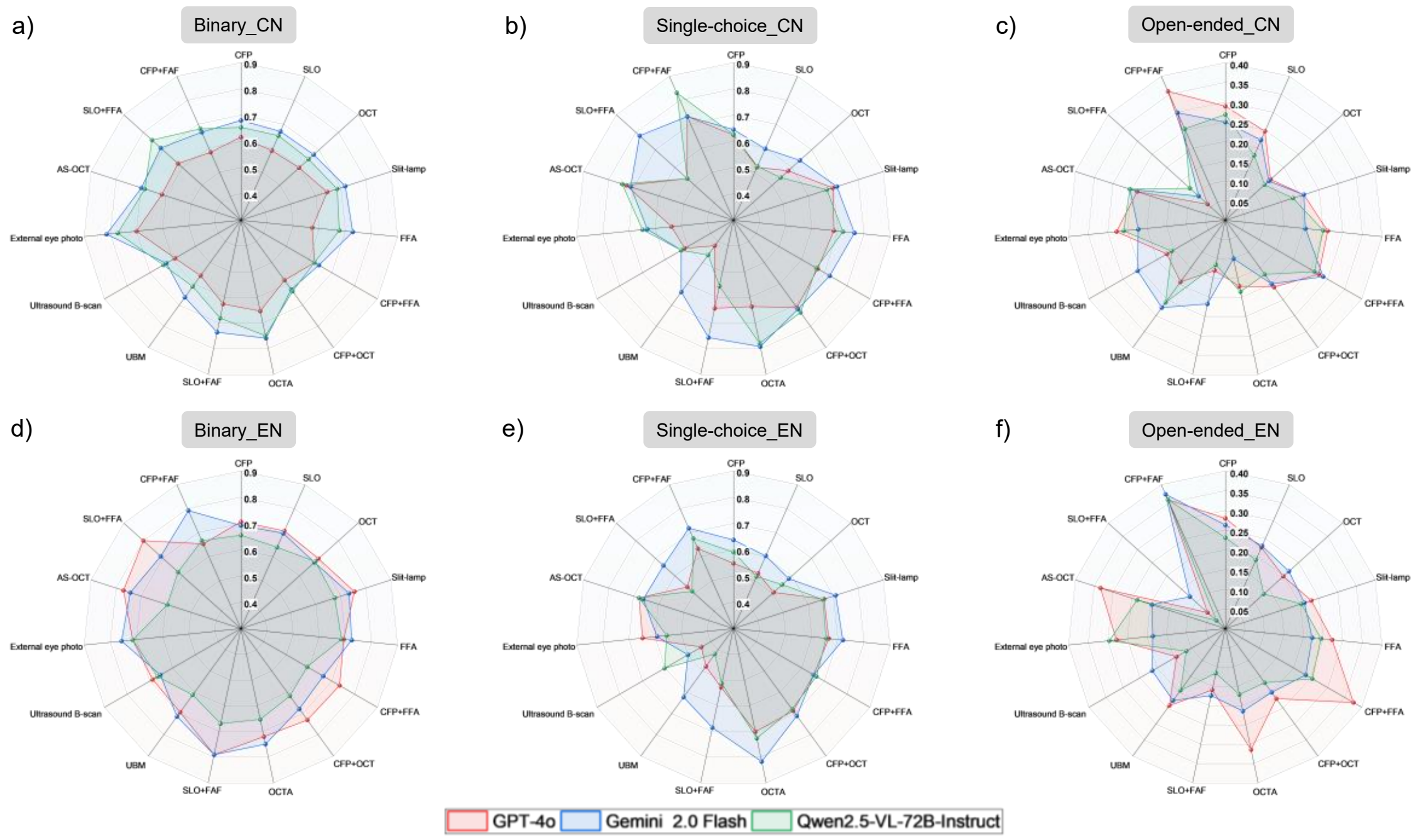

**Supplementary Figure 1. Radar charts illustrating the accuracy of three visual-language models across different question subsets and imaging modalities.** Panels **(a–f)** present model performance in six subsets: **(a)** Binary question-answer (QA) pairs in Chinese (Binary_CN), **(b)** Single-choice QA pairs in Chinese (Single-choice_CN), **(c)** Open-ended QA pairs in Chinese (Open-ended_CN), **(d)** Binary QA pairs in English (Binary_EN), **(e)** Single-choice QA pairs in English (Single-choice_EN), and **(f)** Open-ended QA pairs in English (Open-ended_EN). AS-OCT = anterior segment optical coherence tomography, CFP = color fundus photography, FAF = fundus autofluorescence, FFA = fundus fluorescein angiography, OCT = optical coherence tomography, OCTA = optical coherence tomography angiography, SLO = scanning laser ophthalmoscopy, UBM = ultrasound biomicroscopy.

**Supplementary Table 1 | Prompt used for QA pairs generation.**

| | **Prompt** |
|---|---|
| System prompt | You are an assistant specialized in generating question-answer (QA) pairs based solely on the provided text related to image lesions and diagnosis. |
| User prompt for binary QA | Ask True or False questions related to image lesions and diagnosis described in the provided text.<br>The QA pairs must strictly comply with the following rules:<br>1. The questions should be answerable with the information provided in the caption, and the correctness of the questions should be approximately balanced between True and False.<br>2. **Focus only on the content of the images and their diagnosis.** Do not include any information about ages, genders, symptoms, medical history, visual acuity, treatment, risk, significance, or management in questions or answers.<br>3. If the text describes **multiple images**, generate separate QA pairs for each image, ensuring clarity and relevance.<br>4. If the text only mentions the **name of a disease**, treat it as the diagnostic result, and only generate QA pairs about the diagnosis itself.<br>5. All QA pairs must be **non-repetitive**, derived solely from the text, and must not contain any information not explicitly described in the text.<br>6. Generate each QA pair in **both English and Chinese**, ensuring the questions and answers are concise, accurate, and consistent with the text.<br>7. **Avoid including any information unrelated to the images or lesions**, such as inferred causes, patient demographics, or treatment outcomes.<br>8. Use the following output format for each QA pair:<br>Question: Generated English question, Answer: True/False.<br>问：生成的中文问题：答：对/错。 |
| User prompt for single-choice QA | Ask questions related to image lesions and diagnosis described in the provided text and generate four options for each question.<br>The QA pairs must strictly comply with the following rules:<br>1. The questions should be answerable with the information provided in the caption, and the four options should include one correct and three incorrect options, with the position of the correct option randomized.<br>2. **Focus only on the content of the images and their diagnosis.** Do not include any information about ages, genders, symptoms, medical history, visual acuity, treatment, risk, significance, or management in questions or answers.<br>3. If the text describes **multiple images**, generate separate QA pairs for each image, ensuring clarity and relevance.<br>4. If the text only mentions the **name of a disease**, treat it as the diagnostic result, and only generate QA pairs about the diagnosis itself.<br>5. All QA pairs must be **non-repetitive**, derived solely from the text, and must not contain any information not explicitly described in the text.<br>6. Generate each QA pair in **both English and Chinese**, ensuring the questions and answers are concise, accurate, and consistent with the text.<br>7. **Avoid including any information unrelated to the images or lesions**, such as inferred causes, patient demographics, or treatment outcomes.<br>8. Use the following output format for each QA pair:<br>Question: Generated English question and question’ choice: A: option content B: option content C: option content D: option content, Answer: The correct option(A\B\C\D).<br>问：生成的中文问题和问题选项：A：选项内容 B：选项内容 C：选项内容 D：选项内容，答：正确选项（A\B\C\D）。 |
| User prompt for open-ended QA | Your task is to generate **open-ended question-answer (QA) pairs** specifically related to image lesions and diagnosis described in the provided text. The QA pairs must strictly comply with the following rules:<br>1. **Focus only on the content of the images and their diagnosis.** Do not include any information about ages, genders, symptoms, medical history, visual acuity, treatment, risk, significance, or management in questions or answers.<br>2. If the text describes **multiple images**, generate separate QA pairs for each image, ensuring clarity and relevance.<br>3. If the text only mentions the **name of a disease**, treat it as the diagnostic result, and only generate QA pairs about the diagnosis itself.<br>4. All QA pairs must be **non-repetitive**, derived solely from the text, and must not contain any information not explicitly described in the text.<br>5. Generate each QA pair in **both English and Chinese**, ensuring the questions and answers are concise, accurate, and consistent with the text.<br>6. **Avoid including any information unrelated to the images or lesions**, such as inferred causes, patient demographics, or treatment outcomes.<br>7. Use the following output format for each QA pair:<br>Q: Generated English question, A: Generated English answer.<br>问：生成的中文问题，答：生成的中文答案。 |

**Supplementary Table 2 | Keywords used to exclude QA pairs that could not be answered based on image content.**

| | Keywords |
|---|---|
| English keywords | how old\|what is the age\|gender\|is the patient male\|is the patient female\|male or female\|boy or girl\|child or adult\|elderly patient\|infant\|teenager\|complain\|complaint\|blurry vision duration\|how long has vision been blurry\|how long has the patient had photophobia\|complains of tearing\|reports eye pain\|eye soreness\|eye pressure sensation\|foreign body sensation\|has a headache\|feels nauseous\|experiencing diplopia\|experiencing photophobia\|experiencing dry eyes\|experiencing itching\|subjective symptoms\|visual discomfort\|when did symptoms start\|how long have symptoms lasted\|has this happened before\|first episode\|recurrent episodes\|past medical history\|ocular history\|family history\|drug history\|history of eye drops\|history of artificial tears\|past surgeries\|history of laser treatment\|anti-VEGF treatment\|has diabetes\|has hypertension\|has glaucoma\|refractive error\|myopia degree\|hyperopia degree\|astigmatism degree\|BCVA result\|visual acuity test result\|wearing glasses\|contact lens use\|IOP test result |
| Chinese keywords | 几岁\|年龄\|性别\|男孩\|女孩\|男性\|女性\|男人\|女人\|儿童\|成人\|老年人\|婴儿\|青少年\|主诉\|视力下降多久\|视力模糊多久\|看不清楚多久\|视物变形多久\|畏光\|流泪\|疼痛感\|酸胀感\|异物感\|是否头痛\|有无头痛\|是否恶心\|视物旋转\|是否复视\|是否畏光\|有无流泪\|有无疼痛\|是否干涩\|是否眼红\|是否瘙痒\|病史\|既往史\|家族史\|用药史\|服药史\|是否滴眼药水\|是否使用人工泪液\|使用过什么药\|手术史\|是否做过手术\|是否接受过激光治疗\|是否接受过抗 VEGF 治疗\|有无糖尿病史\|有无高血压史\|有无青光眼史\|多长时间\|持续多久\|什么时候开始\|症状持续了多久\|首次发作\|是否复发\|以前是否发生过\|多久以前出现\|是否做过检查\|是否戴隐形眼镜\|是否佩戴眼镜\|近视度数\|远视度数\|散光度数\|视力检查结果 |

**Supplementary Table 3 | Prompt used for balance the TRUE/FALSE (A) and correct answer options (B).**

| | Prompt |
|---|---|
| A | Modify the following text to contain incorrect information based on the given caption and return the altered version.<br>Requirements:<br>1. Change key facts to be incorrect while keeping the text fluent and grammatically correct.<br>2. Ensure the incorrect information remains understandable and appears reasonable.<br>3. Maintain the original language of the text without adding any extra words or greetings.<br>Caption: {caption}<br>Original text: {question} |
| B | Instructions:<br>Please strictly follow the steps below and return only the modified result:<br><br>1. Identify the correct answer option (content: {correct_answer}) and the target option (content: {change_answer}) in the given question and options.<br><br>2. Swap their contents:<br>Replace the correct answer option's content with the target option's content.<br>Replace the target option's content with the correct answer option's content.<br>3. Keep everything else unchanged, including:<br>The rest of the question.<br>The order of all answer choices (A, B, C, D).<br>4. Return only the modified question and all options without any greetings, explanations, or extra text.<br><br>Input:<br>Question and options: {question}<br>Correct answer option: {correct_answer}<br>Target option: {change_answer} |

**Supplementary Table 4 | Prompt used for generating response from large language models.**

| | **Prompt*** |
|---|---|
| A | 请基于图片判断以下文本的内容是否正确，如正确请回答“对”，错误请回答“错”，不允许出现其他回答。<br>文本: {question}。 |
| B | Based on the provided image, judge the text. If the text is correct, respond with 'True'; if it is incorrect, respond with 'False'. No other responses are allowed.<br>Here is the text: {question}. |
| C | 请基于图片回答以下单选题。只能从'A', 'B', 'C', 'D'中选择一个回答，不允许出现其他回答。<br>问题: {question}。 |
| D | Answer the question based on the provided image. Only one of the options 'A', 'B', 'C', or 'D' is allowed; no other responses are permitted.<br>Here is the question: {question}. |
| E | 请基于图片用中文回答问题。回答尽可能准确和简短，不要出现问候语等任何非必要的回答。<br>问题: {question}。 |
| F | Answer the question based on the provided image. Answer it as briefly and accurately as possible, without greetings or other unnecessary responses.<br>Question: {question}. |

*Images were encoded with Base64 format and input with the following prompts and question.

**Supplementary Table 5 | Prompt used to determine the correctness of responses to open-ended questions in Chinese (A) and English (B).**

<table>
<tr><th></th><th>Prompt</th></tr>
<tr><td>A</td><td>请根据以下标准判断给定答案是否正确：<br>1. 如果给定答案与预期答案在事实和语义上保持一致，则回答 TRUE，否则回答 FALSE。<br>2. 忽略语言表达上的细微差异，只关注核心事实是否正确。<br>3. 如遇模棱两可或信息不完整的情况，依据常识和上下文做出合理判断。<br>4. 你的最终回答只能是 TRUE 或 FALSE，不得包含其他内容。<br><br>问题: {question}<br>预期答案: {expected}<br>给定答案: {given}<br><br>请判断给定答案是否正确：</td></tr>
<tr><td>B</td><td>Evaluate whether the given answer is correct based on the following criteria:<br><br>If the given answer aligns with the expected answer in both factual accuracy and meaning, respond with TRUE; otherwise, respond with FALSE.<br>Ignore minor differences in wording and focus only on the correctness of the core facts.<br>If the answer is ambiguous or incomplete, make a reasonable judgment based on common sense and context.<br>Your final response must be either TRUE or FALSE, with no additional text.<br>Question: {question}<br>Expected Answer: {expected}<br>Given Answer: {given}<br><br>Is the given answer correct?</td></tr>
</table>

**Supplementary Table 6 | Modality frequency of the images used in this study.**

| Modality | Frequency |
|---|---|
| CFP | 689 |
| SLO | 652 |
| OCT | 645 |
| Slit-lamp | 550 |
| FFA | 130 |
| CFP+FFA | 65 |
| CFP+OCT | 64 |
| OCTA | 62 |
| SLO+FAF | 61 |
| UBM | 56 |
| Ultrasound B-scan | 55 |
| External eye photo | 49 |
| AS-OCT | 40 |
| SLO+FFA | 28 |
| CFP+FAF | 28 |
| Visual Field | 26 |
| CT | 18 |
| Corneal confocal | 17 |
| FAF | 17 |
| IR | 17 |
| Retro mode | 15 |
| ICGA | 14 |
| SLO+OCT | 10 |
| CFP+FFA+ICGA | 9 |
| CFP+IR | 9 |
| CFP+FFA+OCT | 9 |
| CFP+ICGA | 7 |
| MRI | 6 |
| SLO+FAF+FFA | 6 |
| Retcam | 6 |
| CFP+IR+Red-free | 5 |
| FFA+ICGA | 5 |
| IOLMaster | 4 |
| Red-channel | 4 |
| CFP+FAF+OCT | 4 |
| CFP+FAF+FFA | 4 |
| CFP+FAF+Ultrasound B-scan | 3 |
| Gonioscope | 3 |
| Slitlamp+CFP | 3 |
| External eye photo + Slitlamp | 3 |
| Corneal topography | 3 |
| OCT+OCTA | 2 |
| Meibography | 2 |
| CFP+Slitlamp | 2 |
| FFA+OCT | 2 |
| External eye photo + MRI + SS | 2 |
| CFP+FAF+FFA+ICGA | 2 |
| Red-free | 2 |
| CFP+FFA+Ultrasound B-scan | 2 |
| Slitlamp+AS-OCT | 2 |
| SLO+Ultrasound B-scan | 2 |
| CFP+OCT+OCTA | 2 |
| CFP+Red-free+OCT | 2 |
| CFP+IR+FFA+ICGA | 1 |
| Ultrasound A-scan + B-scan | 1 |
| AS-OCT+PS-OCT | 1 |
| Slitlamp+OCTA | 1 |
| iTrace | 1 |
| CFP+Red-channel+Blue-channel+Green-channel+IR+OCT | 1 |
| IR+FFA+Ultrasound B-scan+OCT | 1 |
| CFP+Red-channel+Blue-channel+Green-channel | 1 |
| External eye photo + MRI | 1 |
| Slitlamp+Corneal topography | 1 |
| CFP+Red-channel+Blue-channel+Green-channel+OCT | 1 |
| IR+FAF | 1 |
| CFP+IR+Red-free+OCT | 1 |
| IR+OCT | 1 |
| UBM+Retcam | 1 |
| Slitlamp + Ultrasound B-scan | 1 |
| IR+CFP+FFA | 1 |
| CFP+FAF+IR | 1 |
| Blue-channel | 1 |
| Green-channel | 1 |
| ICGA+OCT | 1 |
| Slitlamp+OCT | 1 |
| FFA+Ultrasound B-scan | 1 |
| Slitlamp+UBM | 1 |
| VEP | 1 |
| CFP+SLO | 1 |
| Slitlamp+Ultrasound B-scan+CFP+MRI | 1 |
| Red-free+MRI | 1 |
| CFP+OCT+Ultrasound B-scan | 1 |
| CFP+Ultrasound B-scan+FFA+OCT+FAF | 1 |
| Slitlamp+OCT+CFP+Ultrasound B-scan | 1 |
| SLO+FFA+OCTA | 1 |
| Slitlamp+OCT+CFP+FFA | 1 |
| External eye photo + CT + SS | 1 |
| External eye photo + CFP + FFA | 1 |
| OCT+Ultrasound B-scan | 1 |
| IR+FFA+Ultrasound B-scan | 1 |
| Slitlamp+FFA | 1 |
| CFP+FAF+FFA+OCT | 1 |
| SLO+OCT+FFA+ICGA | 1 |
| CFP+FAF+ICGA | 1 |
| SLO+OCT+FFA | 1 |
| SLO+FAF+ICGA | 1 |
| CFP+Ultrasound B-scan | 1 |
| Total | 3469 |

**Supplementary Table 7 | Comparison of OphthalWeChat with Existing Ophthalmic VQA Datasets**

| Dataset | Ophthalmic Modalities | Number of images | Number of QA pairs | Close-ended QA | Open-ended QA | Multilanguage | Image Source |
|---|---|---|---|---|---|---|---|
| **Diabetic Macular Edema VQA Dataset[22]** | 1 (CFP) | 689 | 13,470 | Yes | No | No | Public datasets |
| **FunBench[19]** | 3 (CFP, OCT, UWF) | 16,348 | 91,810 | Yes | No | No | Public datasets |
| **GMAI-MMBench*[23]** | 6 (CFP, OCT, CT, IR, Adaptive Optics Ophthalmoscopy, Microscopy) | 2,450 | - | Yes | No | No | Public sources and hospitals |
| **Liang et al., 2025[20]** | 2 (CFP and OCT) | 514 | 514 | Yes | No | No | Public datasets |
| **OmniMedVQA*[24]** | 2 (CFP and OCT) | 13,899 | 15,461 | Yes | No | No | Public datasets |
| **OphthalVQA[21]** | 6 (CFP, SLO, OCT, FFA, Slit-lamp, Ultrasound) | 60 | 600 | No | Yes | No | Private dataset |
| **PMC-VQA*[15]** | ＞5 (e.g., CFP, SLO, OCT, Slit-lamp, FFA, etc.) | - | - | Yes | Yes | No | PubMedCentral (PMC) |
| **OphthalWeChat (ours)** | 29 (e.g., CFP, SLO, OCT, Slit-lamp, FFA, etc.) | 3,469 | 30,120 | Yes | Yes | Yes | Official WeChat Accounts |

*For datasets involving multiple anatomical regions, only the ophthalmology-related data are included in this analysis. CFP = Color Fundus Photography; OCT = Optical Coherence Tomography; FFA = Fundus Fluorescein Angiography; ICGA = Indocyanine Green Angiography; IR = Infrared Reflectance; UWF = Ultrawide-field Fundus Imaging; QA = Question Answering; SLO = Scanning Laser Ophthalmoscopy; VQA = Visual Question Answering.